\PassOptionsToPackage{bookmarks=false}{hyperref}
\documentclass[sigconf]{acmart} 
\usepackage{xcolor}
\usepackage{color}
\usepackage{amsmath}
\usepackage{algorithmicx}
\usepackage{url}
\usepackage{bm}
\usepackage{listings}
\usepackage{multirow}
\usepackage{todonotes}
\usepackage{subfig}
\usepackage{listing}
\usepackage{arydshln}
\usepackage{tikz}
\usepackage{pifont}
\usetikzlibrary{tikzmark}
\usepackage{wrapfig}
\usepackage{balance}
\usepackage{enumitem}
\usepackage{booktabs}
\setlist{leftmargin=3mm}

\usepackage{xspace}
\newcommand{\ie}{{\emph{i.e.}},\xspace}

\newcommand{\eg}{{\emph{e.g.}},\xspace}
\newcommand{\etc}{\emph{etc.}\xspace}

\newcommand{\GH}{{\sc GitHub}\xspace}

\newcommand{\SO}{{\sc Stack Overflow}\xspace}

\definecolor{deepblue}{rgb}{0,0,0.5}
\definecolor{deepred}{rgb}{0.6,0,0}
\definecolor{deepgreen}{rgb}{0,0.5,0}
\definecolor{darkgreen}{RGB}{43,163,39}

\newcommand{\cmark}{\textcolor{darkgreen}{\ding{51}}}
\newcommand{\xmark}{\textcolor{red}{\ding{55}}}

\lstdefinestyle{python}{
  language=Python,
  otherkeywords={self},
  keywordstyle=\bfseries\color{deepblue},
  emph={MyClass,__init__},
  emphstyle=\color{deepred},
  showstringspaces=false,
  breaklines=true,
  columns=fullflexible,
  escapeinside=||,
  basicstyle=\fontfamily{cmtt}\footnotesize,
  aboveskip=-0.6\baselineskip
}

\lstdefinestyle{java}{
  language=Java,
  otherkeywords={class, public},
  keywordstyle=\bfseries\color{deepblue},
  emph={MyClass,__init__},
  emphstyle=\color{deepred},
  showstringspaces=false,
  breaklines=true,
  columns=fullflexible,
  escapeinside=||,
  basicstyle=\fontfamily{cmtt}\footnotesize,
  aboveskip=-0.6\baselineskip
}

\lstdefinestyle{python_standalone}{
  language=Python,
  otherkeywords={self},
  keywordstyle=\bfseries\color{deepblue},
  emph={MyClass,__init__},
  emphstyle=\color{deepred},
  showstringspaces=false,
  breaklines=true,
  columns=fullflexible,
  escapeinside=||,
  basicstyle=\fontfamily{cmtt}\footnotesize,
  frame=none
}

\makeatletter
\g@addto@macro\normalsize{%
  \setlength\abovedisplayskip{2pt}
  \setlength\belowdisplayskip{2pt}
  \setlength\abovedisplayshortskip{2pt}
  \setlength\belowdisplayshortskip{2pt}
}
\makeatother

\setcopyright{acmcopyright}

\copyrightyear{2018} 
\acmYear{2018} 
\setcopyright{acmcopyright}
\acmConference[MSR '18]{MSR '18: 15th International Conference on Mining Software Repositories }{May 28--29, 2018}{Gothenburg, Sweden}
\acmPrice{15.00}
\acmDOI{10.1145/3196398.3196408}
\acmISBN{978-1-4503-5716-6/18/05}

\newcommand\blfootnote[1]{%
  \begingroup
  \renewcommand\thefootnote{}\footnote{#1}%
  \addtocounter{footnote}{-1}%
  \endgroup
}

\begin{document}
\title[Mining Aligned NL-Code Pairs from Stack Overflow]{Learning to Mine Aligned Code and \\ Natural Language Pairs from Stack Overflow}

\author{Pengcheng Yin,$^*$ ~Bowen Deng,$^*$ ~Edgar Chen, ~Bogdan Vasilescu, ~Graham Neubig}
\affiliation{%
  \institution{Carnegie Mellon University, USA}
  \streetaddress{Carnegie Mellon University, USA}
  \city{\{pcyin, bdeng1, edgarc, bogdanv, gneubig\}@cs.cmu.edu}
}

\renewcommand{\shortauthors}{P.\ Yin, B.\ Deng, E.\ Chen, B.\ Vasilescu, G.\ Neubig}


\begin{abstract}
For tasks like code synthesis from natural language, code retrieval, and code
summarization, data-driven models have shown great promise. However, creating
these models require parallel data between natural language (NL) and code with
fine-grained alignments. \SO (SO) is a promising source to create
such a data set: the questions are diverse and most of them have corresponding
answers with high quality code snippets. However, existing heuristic methods
(\eg pairing the title of a post with the code in the accepted answer) are
limited both in their coverage and the correctness of the NL-code pairs obtained. In
this paper, we propose a novel method to mine high-quality
aligned data from SO using two sets of features: hand-crafted features considering
the structure of the extracted snippets, and correspondence features
obtained by training a probabilistic model to capture the correlation between
NL and code using neural networks.
These features are fed into a classifier that determines the quality of mined
NL-code pairs.
Experiments using Python and Java as test beds show that the proposed
method greatly expands coverage and accuracy over existing
mining methods, even when using only a small number of labeled examples.
Further, we find that reasonable results are achieved even when training the classifier
on one language and testing on another, showing promise for scaling NL-code
mining to a wide variety of programming languages beyond those for which we
are able to annotate data.
\end{abstract}

\maketitle

%

\section{Introduction}

Recent years 
\blfootnote{$^*$ PY and BD contributed equally to this work.}
have witnessed the burgeoning of a new suite of developer assistance
tools based on natural language processing (NLP) techniques, 
for code completion~\cite{franks2015cacheca}, source code
summarization~\cite{allamanis2016convolutional}, automatic documentation of source
code~\cite{wong2013autocomment}, 
deobfuscation~\cite{raychev2015predicting, vasilescu2017jsnaughty, decompiled-names}, cross-language
porting~\cite{nguyen2013lexical,nguyen2014statistical}, code 
retrieval~\cite{wei2015building,allamanis2015bimodal} and even code synthesis 
from natural language~\cite{quirk2015language, desai2016program, locascio2016regex,
yin2017acl}.

Besides the creativity and diligence of the researchers involved, these recent 
success stories can be attributed to two properties of software source code. 
\emph{First}, it is highly repetitive~\cite{gabel2010study,devanbu2015new}, 
therefore predictable in a statistical sense. 
This statistical predictability enabled researchers to expand from models of source 
code and natural language (NL) created using hand-crafted rules, which have a 
long history~\cite{miller1981natural}, to data-driven models that have proven flexible, 
relatively easy-to-create, and often more effective than corresponding hand-crafted
precursors~\cite{hindle2016naturalness,nguyen2014statistical}.
\emph{Second}, source code is available in large amounts, thanks to the proliferation 
of open source software in general, and the popularity of open access, ``Big Code'' 
repositories like \GH and \SO (SO); these platforms host tens of millions 
of code repositories and programming-related questions and answers, respectively, 
and are ripe with data that can, and is, being used to train such models \cite{raychev2015predicting}.

However, the statistical models that power many such applications are only as 
useful as the data they are trained on, \ie garbage in, garbage out~\cite{sheng2008get}.
For a particular class of applications, such as source code retrieval given a NL
query~\cite{wei2015building}, 
source code summarization in NL~\cite{iyer2016summarizing}, 
and source code synthesis from NL~\cite{yin2017acl,rabinovich17syntaxnet}, 
all of which involve correspondence between NL utterances and code, it is essential to have 
access to \emph{high volume, high quality, parallel data}, in which NL and source 
code align closely to each other.

\begin{figure}[!t]
\centering
\includegraphics[width=\columnwidth, clip=true]{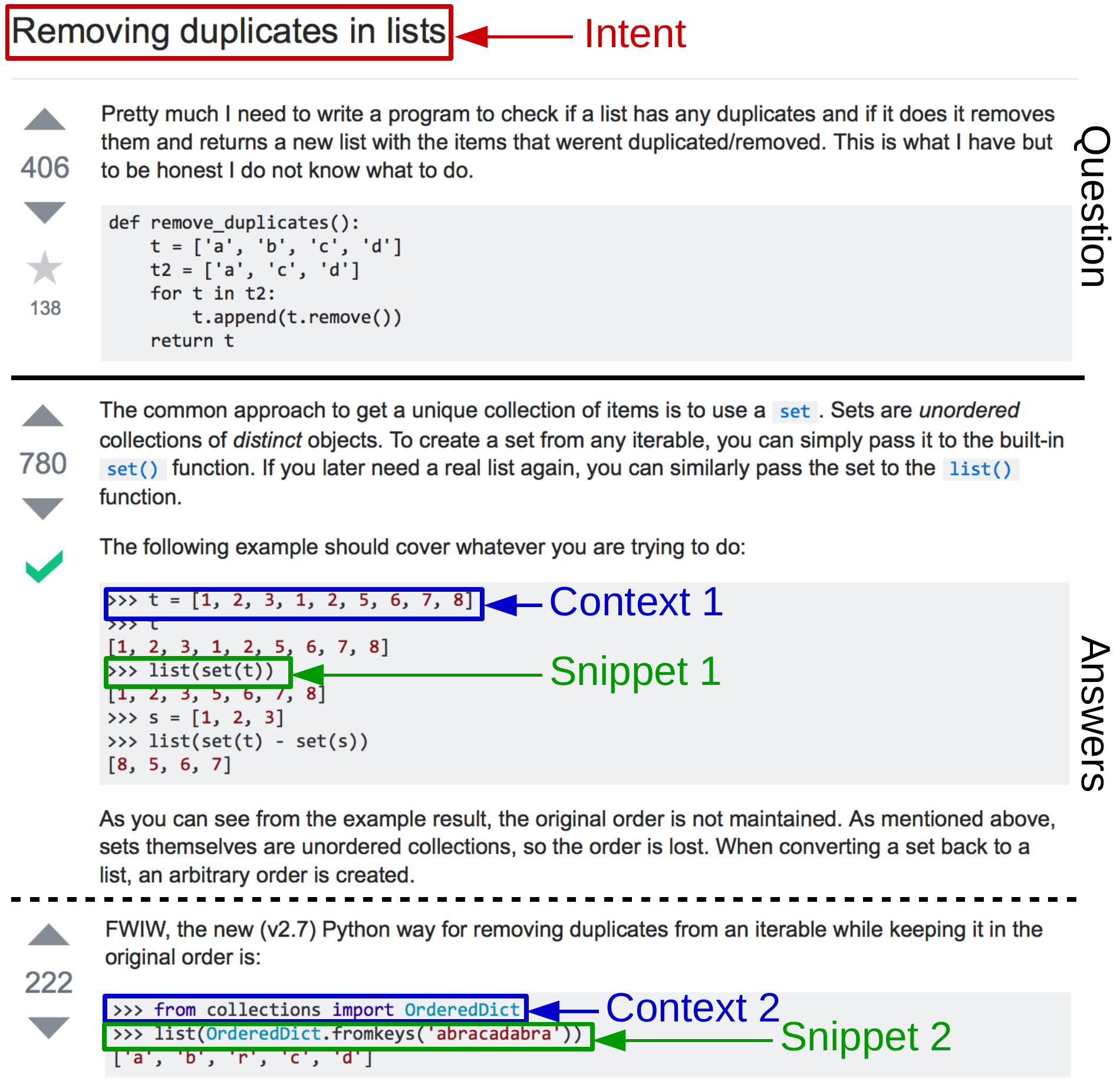}
\caption{Excerpt from a SO post showing two answers, and the corresponding NL 
intent and code pairs.\vspace{-0.2cm}}
\label{fig:example}
\end{figure}

While one can hope to mine such data from Big Code repositories like 
SO, straightforward mining approaches may also extract quite a bit of noise.
We illustrate the challenges associated with mining aligned (parallel) pairs 
of NL and code from SO with the example of a Python 
question in Figure~\ref{fig:example}.
Given a NL query (or intent), 
\eg 
``removing duplicates in lists'', and the goal of finding its matching source code snippets
among the different answers, 
prior work used either a straightforward mining approach that simply picks all 
code blocks that appear in the answers~\cite{allamanis2015bimodal},
or one that picks all code blocks from answers that are 
highly ranked or \emph{accepted}~\cite{iyer2016summarizing, wong2013autocomment}.\footnote{
There is at most one accepted answer per question; see green check symbol in Fig~\ref{fig:example}.}
However, it is not necessarily the case that \emph{every} code block accurately reflects the intent.
Nor is it that the \emph{entire} code block is answering
the question; some parts may simply describe the context, such as variable definitions (Context 1) or
import statements (Context 2), while other parts might be entirely irrelevant (\eg the
latter part of the first code block).

There is an inherent trade-off here between scale and data quality. 
On the one hand, when mining pairs of NL and code from SO, one could devise filters 
using features of the SO questions, answers, and the specific programming language 
(\eg only consider accepted answers with a single code block or with high vote counts,
or filtering out \texttt{print} statements in Python, much like one thrust of prior work~\cite{wong2013autocomment,iyer2016summarizing}); 
fine-tuning heuristics may achieve high pair quality, but this inherently reduces the
size of the mined data set and it may also be very language-specific.
On the other hand, extracting all available code blocks, much like the other thrust of
prior work~\cite{allamanis2015bimodal}, scales better but adds noise (and still cannot 
handle cases where the ``best'' code snippets are smaller than a full code block).
Ideally, a mining approach to extract parallel pairs would handle these tricky cases
and would operate at scale, 
extracting \emph{many high-quality pairs}.
To date, none of the prior work approaches satisfies both requirements of high quality
and large quantity.


In this paper, we propose a novel technique that fills this gap 
(see Figure~\ref{fig:overview} for an overview).
Our key idea is to treat the problem as a classification problem: given an NL intent 
(\eg the SO question title) and \emph{all} contiguous code fragments 
extracted from all answers of that question as candidate matches (for each answer code 
block, we consider all line-contiguous fragments as candidates, \eg for a 
3-line code block 1-2-3, we consider fragments consisting of lines 1, 2, 3, 1-2, 2-3, 
and 1-2-3), we use a 
data-driven classifier to decide if a candidate aligns well with the NL intent.
Our model uses two kinds of information to evaluate candidates: 
(1)~\emph{structural features}, which are hand-crafted but largely language-independent, 
and try to estimate whether a candidate code fragment is valid syntactically, 
and (2)~\emph{correspondence features}, automatically learned, which try to estimate 
whether the NL and code correspond to each other semantically. 
Specifically, for the latter we use a model inspired by recent developments in neural network 
models for machine translation~\cite{bahdanau2015alignandtranslate}, which can calculate 
bidirectional conditional probabilities of the code given the NL and vice-versa.
We evaluate our method on two small labeled data sets of 
Python and Java code that we created from SO.
We show that our approach can extract significantly more, and significantly more accurate 
code snippets in both languages than previous baseline approaches.
We also demonstrate that the classifier is still effective even when trained on Python then 
used to extract snippets for Java, and vice-versa, which demonstrates potential for 
generalizability to other programming languages without laborious annotation of correct 
NL-code pairs.

Our approach strikes a good balance between training effort, scale, and accuracy: 
the correspondence features can be trained without human intervention on readily 
available data from SO; the structural features are simple and easy to apply to new 
programming languages; and the classifier requires minimal amounts of manually 
labeled data (we only used 152 Python and 102 Java manually-annotated SO question threads in total).
Even so, compared to the heuristic techniques from prior work~\cite{allamanis2015bimodal,
wong2013autocomment, iyer2016summarizing}, our approach is able to extract up to 
an order of magnitude more aligned pairs with no loss in accuracy, or reduce errors 
by more than half while holding the number of extracted pairs constant.

Specifically, we make the following contributions:
\begin{itemize}
\item We propose a novel technique for extracting aligned NL-code pairs from SO posts,
based on a classifier that combines snippet structural features, readily extractable, with bidirectional conditional probabilities, estimated using a state-of-the-art neural network model for machine translation.
\item We propose a protocol and tooling infrastructure for generating labeled training data.
\item We evaluate our technique on two data sets for Python and Java and discuss performance, potential for generalizability to other languages, and lessons learned.
\item All annotated data, the code for the annotation interface and the mining algorithm are available at~\url{http://conala-corpus.github.io}.
\end{itemize}

\section{Problem Setting}
\label{sec:setting}

\SO (SO) is the most popular Q\&A site for programming related questions, home to millions
of users.
An example of the SO interface is shown in Figure~\ref{fig:example}, with a question (in the 
upper half) and a number of answers by different SO users.
Questions can be about anything programming-related, including features of the programming 
language or best practices.
Notably, many questions are of the \textit{``how to''} variety, \ie questions that ask 
how to achieve a particular goal such as \textit{``sorting a list''}, \textit{``merging two dictionaries''}, 
or \textit{``removing duplicates in lists''} (as shown in the example);
for example, around 36\% of the Python-tagged questions are in this category, as
discussed later in Section~\ref{sec:annotation:dataset}.
These \textit{how-to} questions are the type that we focus on in this work, since they are likely to 
have corresponding snippets and they mimic NL-to-code (or vice versa) queries that users 
might naturally make in the applications we seek to enable, \eg code retrieval and synthesis.

Specifically, we focus on extracting triples of three specific elements of the content 
included in SO posts: 
\begin{itemize}
\item \textbf{Intent:} A description in English of what the questioner wants to do; usually 
corresponds to some portion of the post title.
\item \textbf{Context:} A piece of code that does not implement the intent, but is necessary 
setup, \eg import statements, variable definitions.
\item \textbf{Snippet:} A piece of code that actually implements the intent.
\end{itemize}

An example of these three elements is shown in Figure~\ref{fig:example}.
Several interesting points can be gleamed from this example.
\textit{First,} and most important, we can see that not all snippets in the post are implementing 
the original poster's intent: only two of four highlighted are actual examples of how to remove 
duplicates in lists, the other two highlighted are context, and others still are examples of 
interpreter output.
If one is to train, \eg a data-driven system for code synthesis from NL, or code retrieval using NL, 
only the snippets, or portions of snippets, that actually implement the user intent should be used.
Thus, we need a mining approach that can distinguish which segments of code are actually 
legitimate implementations, and which can be ignored.
\textit{Second,} we can see that there are often several alternative implementations with different 
trade-offs (\eg the first example is simpler in that it doesn't require additional modules to be 
imported first). 
One would like to be able to extract all of these alternatives, \eg to present them to users in the 
case of code retrieval\footnote{Ideally one would also like to present a description of the trade-offs, 
but mining this information is a challenge beyond the scope of this work.} or, in the case of code 
summarization, see if any occur in the code one is attempting to summarize.

These aspects are challenging even for human annotators, as we illustrate next.

\section{Manual Annotation}
\label{sec:annotation}

To better understand the challenges with automatically mining aligned NL-code snippet pairs 
from SO posts, we manually annotated a set of labeled NL-code pairs.
These also serve as the gold-standard data set for training and evaluation.
Here we describe our annotation method and criteria, salient statistics about the data collected, 
and challenges faced during annotation.




For each target programming language, we first obtained all questions from the official SO 
data dump\footnote{\label{se-dump}Available online at \url{https://archive.org/details/stackexchange}} 
dated March 2017 by filtering questions tagged with that language. 
We then generated the set of questions to annotate by: 
(1) including all top-100 questions ranked by view count; and 
(2) sampling 1,000 questions from the probability distribution generated by their view counts on SO;
we choose this method assuming that more highly-viewed questions are more 
important to consider as we are more likely to come across them in actual applications.
While each question may have any number of answers, we choose to only annotate the top-3 
highest-scoring answers to prevent annotators from potentially spending a long time on a single 
question.

\subsection{Annotation Protocol and Interface}
\label{sec:annotation:method}







\setlength{\columnsep}{5pt}%
\setlength\intextsep{2pt}
\begin{wrapfigure}{r}{0.5\columnwidth}
  \centering
  \includegraphics[width=\linewidth]{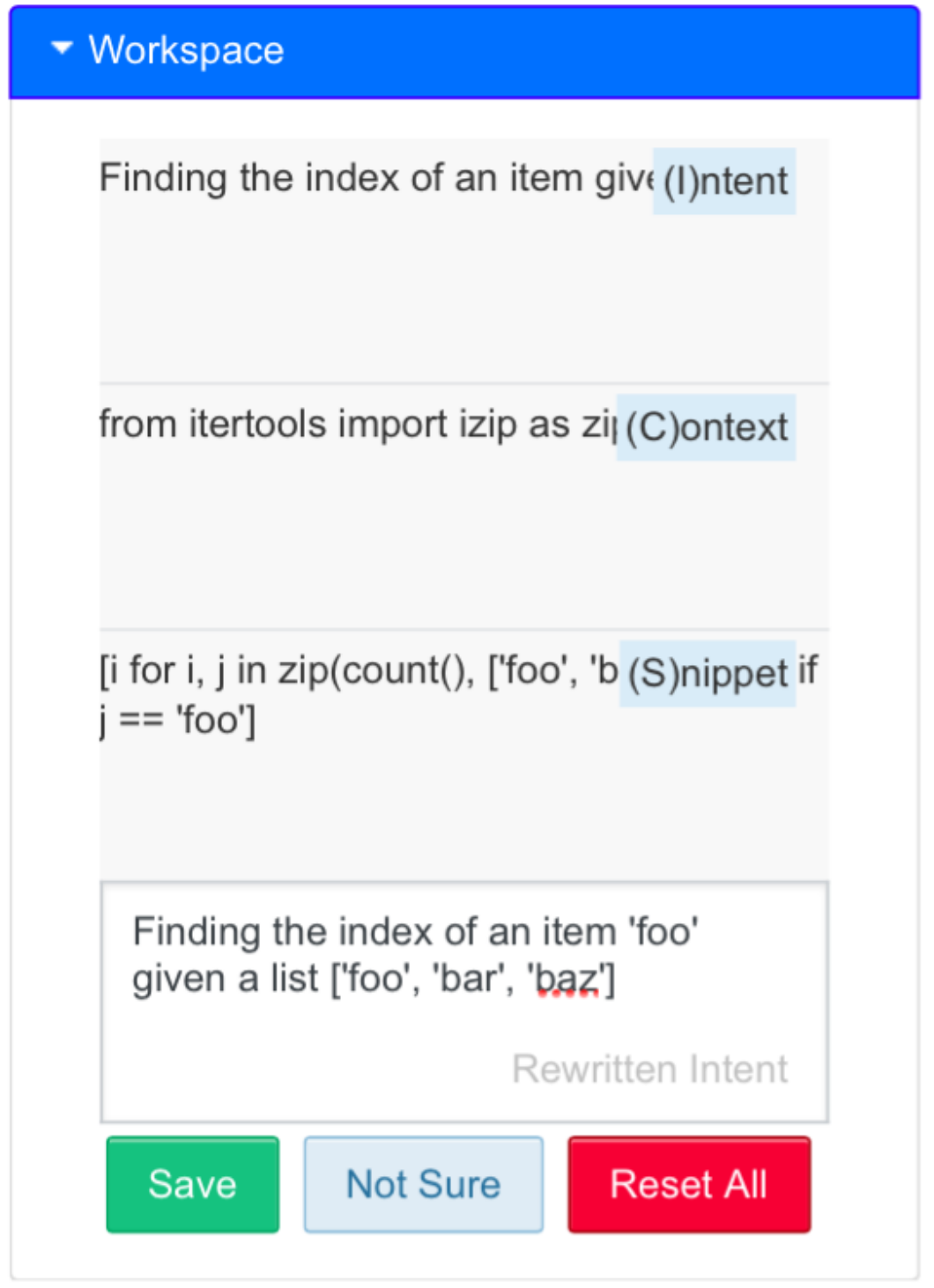}
  \label{fig:nonlin}
  \vskip -1em
\end{wrapfigure}
Consistently annotating the intent, context, and snippet for a variety of posts is not an easy 
task, and in order to do so we developed and iteratively refined a web annotation interface 
and a protocol with detailed annotation criteria and instructions.

The annotation interface allows users to select and label parts of SO posts as (I)intent, (C)ontext, 
and (S)nippet using shortcut keys, as well as rewrite the intent to better match the code 
(\eg adding variable names from the snippet into the original intent), in consideration of 
potential future applications that may require more precisely aligned NL-code data;
in the following experiments we solely consider the intent and snippet, and reserve 
examination of the context and re-written intent for future work.
Multiple NL-code pairs that are part of the same post can be annotated this way.
There is also a ``not applicable'' button that allows users to skip posts that are not of the 
``how to'' variety, and a ``not sure'' button, which can be used when the annotator is uncertain.

The annotation criteria were developed by having all authors attempt to perform 
annotations of sample data, gradually adding notes of the difficult-to-annotate cases to a 
shared document.
We completed several pilot annotations for a sample of Python questions, 
iteratively discussing among the research team the annotation criteria and the difficult-to-annotate 
cases after each, before finalizing the annotation protocol.
We repeated the process for Java posts.
Once we converged on the final annotation standards in both languages, we discarded all pilot 
annotations, and one of the authors (a graduate-level NLP researcher and experienced 
programmer) re-annotated the entire data set according to this protocol.

While we cannot reflect all difficult cases here for lack of space, below is a representative 
sample from the Python instructions:
\begin{itemize}
\item \textbf{Intents:} 
Annotate the command form when possible (\eg ``how do I merge dictionaries'' will be annotated 
as ``merge dictionaries''). 
Extraneous words such as ``in Python'' can be ignored. 
Intents will almost always be in the title of the post, but intents expressed elsewhere that are 
different from the title can also be annotated.
\item \textbf{Context:} 
Contexts are a set of statements that do not directly reflect the annotated intent, but may be 
necessary in order to get the code to run, and include import statements, 
variable definitions, and anything else that is necessary to make sure that the code executes. 
When no context exists in the post this field can be left blank.
\item \textbf{Snippet:}
Try to annotate full lines when possible.
Some special tokens such as ``\texttt{>>>}'', ``\texttt{print}'', and ``\texttt{In[...]}'' that appear at 
the beginning of lines due to copy-pasting can be included.
When the required code is encapsulated in a function, the function definition can be skipped.
\item \textbf{Re-written intent:}
Try to be accurate, but try to make the minimal number of changes to the original intent.
Try to reflect all of the free variables in the snippet to be conducive to future automatic matching 
of these free variables to the corresponding position in code.
When referencing string literals or numbers, try to write exactly as written in the code, and 
surround variables with a grave accent ```''.
\end{itemize}

\setlength{\tabcolsep}{2pt}
\begin{table}[t]
\caption{Details of the labeled data set.} 
\centering
\footnotesize
\begin{tabular}{l|p{.8cm}|p{0.7cm}|p{0.9cm}|p{0.7cm}|p{1.2cm}|p{0.7cm}|p{1.3cm}}
\hline
Lang. & \#Annot. & \#Ques.\ & \#Answer Posts & \#Code Blocks & Avg. Code Length & \%Full Blocks & \%Annot. with Context  \\ \hline 
Python & 527 & 142 & 412 & 736 & 13.2 & 30.7\% & 36.8\% \\ 
Java & 330 & 100 & 297 & 434 & 30.6 & 53.6\% & 42.4\% \\ \hline
\end{tabular}\vspace{-0.1cm}
\label{tab:annotation:stat}
\end{table}

\subsection{Annotation Outcome}
\label{sec:annotation:dataset}


We annotated a total of 418 Python questions and 200 Java questions.
Of those, 152 in Python and 102 in Java were judged as annotatable ({\ie the {\it``how-to''} style 
questions described in Section~\ref{sec:setting}),
resulting in 577 Python and 354 Java annotations.
We then removed the annotations marked as ``not sure'' and all unparsable code snippets.%
\footnote{\label{parsers}We use the built-in \texttt{ast} parser module for Python, and \texttt{JavaParser} for Java.}
In the end we generated 527 Python and 330 Java annotations, respectively.
Table~\ref{tab:annotation:stat} lists basic statistics of the annotations.
Notably, compared to Python, Java code snippets are longer (13.2 vs. 30.6 tokens per snippet), 
and more likely to be full code blocks (30.7\% vs. 53.6\%). 
That is, in close to 70\% of cases for Python, the code snippet best-aligned with the NL intent 
expressed in the question title was \emph{not} a full code block (SO uses special HTML tags
to highlight code blocks, recall the example in Figure~\ref{fig:example}) from one of the answers, 
but rather a subset of it; similarly, the best-aligned Java snippets were \emph{not} full code 
blocks in almost half the cases.
This confirms the importance of mining code snippets beyond the level of entire code blocks,
a limitation of prior approaches.

Overall, we found the annotation process to be non-trivial, which raises several noteworthy 
threats to validity: (1)~it can be difficult for annotators to distinguish between incorrect solutions 
and unusual or bad solutions that are nonetheless correct; (2)~in cases where a single SO 
question elicits many correct answers with many implementations and code blocks, annotators 
may not always label all of them; (3)~long and complex solutions may be mis-annotated; and 
(4)~inline code blocks are harder to recognize than stand-alone code blocks, increasing the risk 
of annotators missing some.
We made a best effort to minimize the impact of these threats by carefully designing and 
iteratively refining our annotation protocol.

\section{Mining Method}
\label{sec:mining}

In this section, we describe our mining method (see Figure~\ref{fig:overview} for an overview).
As mentioned in Section~\ref{sec:setting}, we frame the problem as a classification problem.
First, for every ``how to'' SO question 
we consider its title as the intent
and extract all contiguous lines from across all code blocks
in the question's answers (including 
those we might manually annotate as context; inline code snippets are excluded) as candidate implementations of the intent,
as long as we could parse the candidate snippets.\textsuperscript{\ref{parsers}}
There are some cases where the title is not strictly equal to the intent, which go
beyond the scope of this paper; for the purpose of learning the model we assume the title 
is representative.
This step generates, for every SO question considered, a set of pairs (intent $I$, candidate 
snippet $S$).
For example, the second answer in Figure~\ref{fig:example}, containing a three-line-long 
code block, would generate six line-contiguous candidate snippets, corresponding to 
lines 1, 2, 3, 1-2, 2-3, and 1-2-3.
Our candidate snippet generation approach, though clearly not the only possible approach 
(1)~is simple and language-independent, (2)~is informed by our manual annotations, and 
(3)~it gives good coverage of all possible candidate snippets.

Then, our task is, given a candidate pair ($I$, $S$), to assign a label $y$ representing 
whether or not the snippet $S$ reflects the intent $I$; we define $y$ to equal 1 if the pair 
matches and -1 otherwise.
Our general approach to making this binary decision is to use machine learning to train a 
classifier that predicts, for every pair ($I$, $S$), the probability that $S$ accurately 
implements $I$, \ie $P(y=1|I, S)$, based on a number of \emph{features} 
(Sections~\ref{sec:mining:codestructure} and~\ref{sec:mining:correspondence}).
As is usual in supervised learning, our system first requires an offline \textit{training} 
phase that learns the parameters (\ie feature weights) of the classifier, for which we use the 
annotated data described above (Section~\ref{sec:annotation}).
This way, we can apply our system to an SO page of interest, and compute $P(y=1|I, S)$ 
for each possible intent/candidate snippet pair mined from the SO page. 
We choose logistic regression as our classifier,
as implemented in the \texttt{scikit-learn} Python package. 

As human annotation to generate training data is costly, our goal is to keep the amount of 
manually labeled training data to a minimum, such that scaling our approach to other 
programming languages in the future can be feasible.
Therefore, to ease the burden on the classifier in the face of limited training data, we 
combined two types of features: hand-crafted \emph{structural} features of the code snippets 
(Section~\ref{sec:mining:codestructure}) and machine learned \emph{correspondence} features 
that predict whether intents and code snippets correspond to each-other semantically
(Section~\ref{sec:mining:correspondence}).
Our intuition, again informed by the manual annotation, was that ``good'' and ``bad'' pairs 
can often be distinguished based on simple hand-crafted features; these features could 
eventually be learned (as opposed to hand-crafted), but this would require more labeled 
training data, which is relatively expensive to create.  
 
\begin{figure}[!t]
\centering
\includegraphics[width=\columnwidth, clip=true, trim=70 110 70 110]{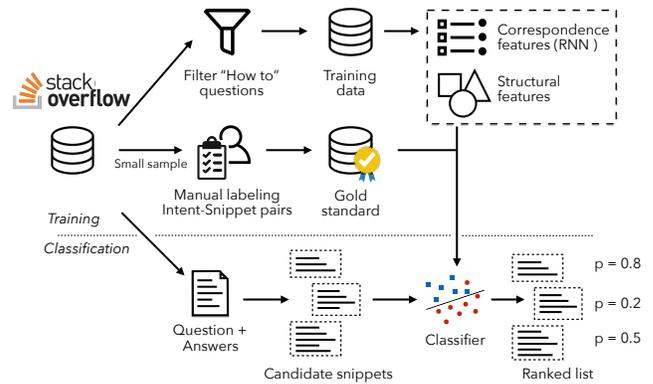}
\caption{Overview of our approach.\vspace{-0.2cm}}
\label{fig:overview}
\end{figure}

\subsection{Hand-crafted Code Structure Features}
\label{sec:mining:codestructure}

The structural features are intended to distinguish whether we can reasonably expect that 
a particular piece of code implements an intent.
We aimed for these features to be both informative and generally applicable to a wide range 
of programming languages. 
These features include the following:
\begin{itemize}
\item \textbf{\textsc{FullBlock}, \textsc{StartOfBlock}, \textsc{EndOfBlock}:}
A code block may represent a single cohesive solution.
By taking only a piece of a code block, we may risk acquiring only a partial solution, and thus we 
use a binary feature to inform the classifier of whether it is looking at a whole code block or not.
On the other hand, as shown in Figure~\ref{fig:example}, many code blocks contain some amount 
of context before the snippet, or other extraneous information, \eg print statements.
To consider these, we also add binary features indicating that a snippet is at the start or 
end of its code block.
\item \textbf{\textsc{ContainsImport}, \textsc{StartsWithAssignment}, \textsc{IsValue}:}
Additionally, some statements are highly indicative of a statement being context or extraneous.
For example, import statements are highly indicative of a particular line being context instead 
of the snippet itself, and thus we add a binary feature indicating whether an import statement 
is included.
Similarly, variable assignments are often context, not the implementation itself, and thus we 
add another feature indicating whether the snippet starts with a variable assignment.
Finally, we observed that in SO (particularly for Python), it was common to have single lines in 
the code block that consisted of only a variable or value, often as an attempt to print these values 
to the interactive terminal.
\item \textbf{\textsc{AcceptedAns}, \textsc{PostRank1}, \textsc{PostRank2}, \textsc{PostRank3}:}
The quality of the post itself is also indicative of whether the answer is likely to be valid or not.
Thus, we add several features indicating whether the snippet appeared in a post that was the 
accepted answer or not, and also the rank of the post within the various answers for a particular 
question.
\item \textbf{\textsc{OnlyBlock}:}
Posts with only a single code block are more likely to have that snippet be a complete 
implementation of the intent, so we added another feature indicating when the extracted snippet 
is the only one in the post.
\item \textbf{\textsc{NumLinesX}:}
Snippets implementing the intent also tend to be concise, so we added features indicating the 
number of lines in the snippet, bucketed into $X=1$, 2, 3, 4-5, 6-10, 11-15, >15.
\item \textbf{Combination Features:}
Some features can be logically combined to express more complex concepts.
E.g., \textsc{AcceptedAns} + \textsc{OnlyBlock} + \textsc{WholeBlock} can express the 
strategy of selecting whole blocks from accepted answers with only one block, as used in 
previous work~\cite{wong2013autocomment,iyer2016summarizing}.
We use this feature and two other combination features: 
specifically $\neg$\textsc{StartWithAssign} + \textsc{EndOfBlock} and 
$\neg$\textsc{StartWithAssign} + \textsc{NumLines1}.
\end{itemize}

\subsection{Unsupervised Correspondence Features}
\label{sec:mining:correspondence}

While all of the features in the previous section help us determine which code snippets are 
likely to implement \textit{some} intent, they say nothing about whether the code snippet 
actually implements \textit{the particular} intent $I$ that is currently under consideration.
Of course considering this correspondence is crucial to accurately mining intent-snippet pairs, 
but how to evaluate this correspondence computationally is non-trivial, as there are very few 
hard and fast rules that indicate whether an intent and snippet are expressing similar meaning.
Thus, in an attempt to capture this correspondence, we take an indirect approach that 
uses a potentially-noisy (\ie not manually validated) but easy-to-construct data set to train 
a probabilistic model to approximately capture 
these correspondences, then incorporate the predictions of this noisily trained model as 
features into our classifier.

\vspace{0.1cm}\emph{Training data of correspondence features:}
Apart from our manually-annotated data set, we collected a relatively large set of intent-snippet 
pairs using simple heuristic rules for learning the correspondence features.
The data set is created by pairing the question titles and code blocks from all SO posts, where 
(1)~the code block comes from an SO answer that was accepted by the original poster, and 
(2)~there is only one code block in this answer.
Of course, many of these code blocks will be noisy in the sense that they contain extraneous 
information (such as extra import statements or variable definitions, \etc), or not directly 
implement the intent at all, but they will still be of use for learning which NL expressions in the 
intent tend to occur with which types of source code.

\vspace{0.1cm}\emph{Learning a model of correspondence:}
Given the training data above, we need to create a model of the correspondence between 
the intent $I$ and snippet $S$.
To this end, we build a statistical model of the bi-directional probability of the intent given the 
snippet $P(I \mid S)$, and the probability of the snippet given the intent $P(S \mid I)$.

Specifically, we follow previous work that has noted that models from \textit{machine translation}  
(MT; \cite{koehn2010smt}) are useful for learning the correspondences between natural language 
and code for the purposes of code summarization~\cite{oda2015learning,iyer2016summarizing}, 
code synthesis from natural language~\cite{locascio2016regex}, and code 
retrieval~\cite{allamanis2015bimodal}.
In particular, we use a model based on \textit{neural MT}~\cite{kalchbrenner2013rctm, 
bahdanau2015alignandtranslate}, a method for MT based on neural networks that is well-suited 
for this task of learning correspondences for a variety of reasons, which we outline below after 
covering the basics.
To take the example of using a neural MT model that attempts to generate an intent $I$ given 
a snippet $S$, these models work by incrementally generating each word of the intent 
$i_1, i_2, \ldots, i_{|I|}$ one word at a time (the exact same process can be performed in the 
reverse direction to generate a snippet $S$ given intent $I$).
For example, if our intent is {\it ``download and save an http file''}, the model would first predict 
and generate {\it ``download''}, then {\it ``and''}, then {\it ``save''}, etc.
This is done in a probabilistic way by calculating the probability of the first word given the snippet 
$P(i_1 \mid S)$ and outputting the word in the vocabulary that maximizes this probability, then 
calculating the probability of the second word given the first word and the snippet 
$P(i_2 \mid S, i_1)$ and similarly outputting the word with the highest probability, etc.
Incidentally, if we already know a particular intent $I$ and want to calculate its probability given 
a particular snippet $S$ (for example to use as features in our classifier), we can also do so by 
calculating the probability of each word in the intent and multiplying them together as follows:
\begin{equation}
\label{eq:probabilitycalc}
P(I \mid S) = P(i_1 \mid S) P(i_2 \mid S, i_1) P(i_3 \mid S, i_1, i_2) \ldots 
\end{equation}

\begin{figure}[!t]
\centering
\includegraphics[width=2.2in]{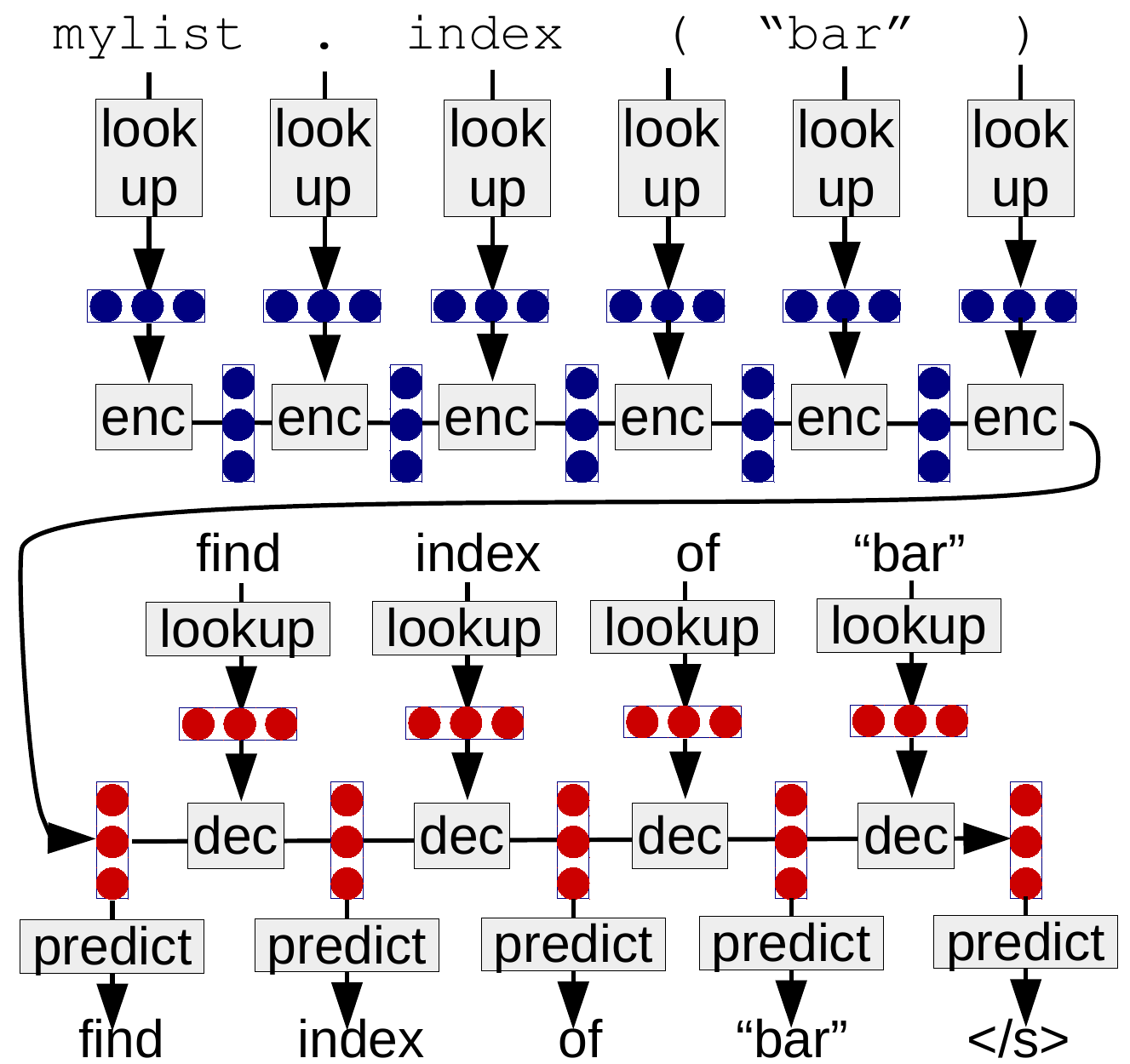}
\caption{An example of neural MT encoder-decoder framework used in calculating 
correspondence scores.}
\label{fig:encdec}
\end{figure}

So how do neural MT models calculate this probability?
We will explain a basic outline of a basic model called the \textit{encoder-decoder 
model}~\cite{sutskever2014sequencetosequence}, and refer readers to references for details \cite{sutskever2014sequencetosequence, bahdanau2015alignandtranslate, neubig2017neural}.
The encoder-decoder model, as shown in Figure~\ref{fig:encdec}, works in two stages: 
First, it \textit{encodes} the input (in this case $S$) into a \textit{hidden vector} of continuous 
numbers $\bm{h}$ using an encoding function
\begin{equation}
\bm{h}_{|S|} = \text{encode}(S).
\end{equation}
This function generally works in two steps: looking up a vector of numbers representing each 
word (often called ``word embeddings'' or ``word vectors''), then incrementally adding 
information about these embeddings one word at a time using a particular variety of network 
called a \textit{recurrent neural network} (RNN).
To take the specific example shown in the figure, at the first time step, we would look up a 
word embedding vector for the first word ``\texttt{mylist}'', $\bm{e}_1 = \bm{e}_{\texttt{mylist}}$ 
and then perform a calculation such as the one below to calculate the hidden vector for the first 
time step:
\begin{equation}
\bm{h}_1 = \text{tanh}(W_{\text{enc,e}} \bm{e}_1 + b_{\text{enc}}),
\end{equation}
where $W_{\text{enc,e}}$ and $b_{\text{enc}}$ are a matrix and vector that are parameters of 
the model, and $tanh(\cdot)$ is the hyperbolic tangent function used to ``squish'' the values to 
be between -1 and 1.
In the next time step, we would do the same for the symbol ``.'', using its embedding 
$\bm{e}_2 = \bm{e}_{\texttt{.}}$, and in the calculation from the second step onward we also 
use the result of the previous calculation (in this case $\bm{h}_2$):
\begin{equation}
\bm{h}_1 = \text{tanh}(W_{\text{enc,h}} \bm{h}_1 + W_{\text{enc,e}} \bm{e}_2 + b_{\text{enc}}).
\label{eq:rnnwithhidden}
\end{equation}
By using the hidden vector from the previous time step, the RNN is able to ``remember'' features 
of the previously occurring words within this vector, and by repeating this process until the end 
of the input sequence, it (theoretically) has the ability to remember the entire content of the input 
within this vector.

Once we have encoded the entire source input, we can use this encoded vector to predict 
the first word of the output.
This is done by multiplying the vector $\bm{h}$ with another weight matrix to calculate a score 
$\bm{g}$ for each word in the output vocabulary:
\begin{equation}
\bm{g}_1 = W_{\text{pred}} \bm{h}_{|S|} + \bm{b}_{\text{pred}}.
\end{equation}
We then predict the actual probability of the first word in the output sentence, for example 
{\it ``find''}, by using the \textit{softmax} function, which exponentiates all of the scores in the 
output vocabulary and then normalizes these scores so that they add up to one:
\begin{equation}
P(i_1 = \text{``find''}) = \frac{\text{exp}(g_\text{find})}{\sum_{\tilde{g}} \text{exp}(\tilde{g})}.
\end{equation}

We use a neural MT model with this basic architecture, with the addition of a feature called 
\emph{attention}, which, put simply, allows the model to ``focus'' on particular words in the 
source snippet $S$ when generating the intent $I$.
The details of attention are beyond the scope of this paper, but interested readers can 
reference \cite{bahdanau2015alignandtranslate, luong-pham-manning:2015:EMNLP}.

\paragraph{Why attentional neural MT models?:}
Attention-based neural MT models are well-suited to the task of learning correspondences 
between natural language intents and code snippets for a number of reasons.
First, they are a purely probabilistic model capable of calculating $P(S \mid I)$ and $P(I \mid S)$, 
which allows them to easily be incorporated as features in our classifier, as described in the 
following paragraph.
Second, they are powerful models that can learn correspondences on a variety of levels; 
from simple phenomena such as direct word-by-word matches~\cite{gu-EtAl:2016:P16-1}, to 
soft paraphrases~\cite{socher2011dynamic}, to weak correspondences between keywords and 
large documents for information retrieval~\cite{huang2013learning}. 
Finally, they have demonstrated success in a number of NL-code related tasks%
~\cite{allamanis2015bimodal, allamanis2016convolutional, locascio2016regex, yin2017acl}, 
which indicates that they could be useful as part of our mining approach as well.

\paragraph{Incorporating correspondence probabilities as features:}
For each intent $I$ and candidate snippet $S$, we calculate the probabilities $P(S \mid I)$ 
and $P(I \mid S)$, and add them as features to our classifier, as we did with the 
hand-crafted structural features in Section~\ref{sec:mining:codestructure}.
\begin{itemize}
\item \textbf{\textsc{SGivenI}, \textsc{IGivenS}:}
Our first set of features are the logarithm of the probabilities mentioned above: 
$\log P(S \mid I)$ and $\log P(I \mid S)$.%
\footnote{We take the logarithm of the probabilities because the actual probability values tend 
to become very small for very long sequences (\eg $10^{-50}$ to $10^{-100}$), while the 
logarithm is in a more manageable range (\eg $-50$ to $-100$).}
Intuitively, these probabilities will be indicative of $S$ and $I$ being a good match because if 
they are not, the probabilities will be low.
If the snippet and the intent are not a match at all, both features will have a low value.
If the snippet and intent are partial matches, but either the snippet $S$ or intent $I$ contain 
extraneous information that cannot be predicted from the counterpart, then \textsc{SGivenI} 
and \textsc{IGivenS} will have low values respectively.
\item \textbf{\textsc{ProbMax}, \textsc{ProbMin}:}
We also represent the max and min of $\log P(S \mid I)$ and $\log P(I \mid S)$.
In particular, the \textsc{ProbMin} feature is intuitively helpful because pairs where the probability 
in \textit{either} direction is low are likely not good pairs, and this feature will be low in the case 
where either probability is low.
\item \textbf{\textsc{NormalizedSGivenI}, \textsc{NormalizedIGivenS}:}
In addition, intuitively we might want the \emph{best} matching NL-code pairs within a particular 
SO page.
In order to capture this intuition, we also normalize the scores over all posts within a particular 
page so that their mean is zero and standard deviation is one (often called the $z$-score).
In this way, the pairs with the best scores within a page will get a score that is significantly 
higher than zero, while the less good scores will get a score close to or below zero.
\end{itemize}

\section{Evaluation}
\label{sec:experiments}

In this section we evaluate our proposed mining approach. 
We first describe the experimental setting in Section~\ref{sec:experiments:settings} before 
addressing the following research questions:
\begin{enumerate}
\item How does our mining method compare with existing approaches across different 
programming languages? (Section~\ref{sec:experiments:results})
\item How do the structural and correspondence features impact the system's performance? 
(Section~\ref{sec:experiments:results})
\item Given that annotation of data for each language is laborious, is it possible 
to use a classifier learned on one programming language to perform mining on other languages? 
(Section~\ref{sec:experiments:transfer_learning})
\item What are the qualitative features of the NL-code pairs that our method succeeds or 
fails at extracting? (Section~\ref{sec:experiments:case_study})
\end{enumerate}

We show that our method clearly outperforms existing approaches and shows potential for reuse \emph{without retraining}, we uncover trade-offs between performance and training complexity, and we discuss limitations, which can inform future work.

\subsection{Experimental Settings}
\label{sec:experiments:settings}

We conduct experimental evaluation on two programming languages: Python and Java.
These languages were chosen due to their large differences in syntax and verbosity, which 
have been shown to effect characteristics of code snippets on SO \cite{yang2016query}.

\vspace{0.1cm}\emph{Learning unsupervised features:}
We start by filtering the SO questions in the Stack Exchange data dump\textsuperscript{\ref{se-dump}} 
by tag (Python and Java), and we use an existing classifier~\cite{iyer2016summarizing} 
to identify the \textit{how-to} style questions.
The classifier is a support vector machine trained by bootstrapping from 100 labeled questions, and achieves over 75\% accuracy as reported in~\cite{iyer2016summarizing}.
We then extract intent/snippet pairs from all these questions as described in Section~\ref{sec:mining:correspondence},
collecting 33,946 pairs for Python and 37,882 for Java.
Next we split the data set into training and validation sets with a ratio of 9:1, keeping the 
90\% for training. 
Statistics of the data set are listed in Table~\ref{tab:exp:unlabeled_data}.%
\footnote{Note that this data may contain some of the posts included in the cross-validation test set with which we evaluate our model later. However, even if it does, we are not using the annotations themselves in the training of the correspondence features, so this does not pose a problem with our experimental setting.}

We implement our neural correspondence model using the DyNet neural network toolkit~\cite{dynet}.
The dimensionality of word embedding and RNN hidden states is 256 and 512.
We use dropout~\cite{Srivastava14dropout}, a standard method to prevent overfitting, on 
the input of the last softmax layer over target words ($p=0.5$), and 
recurrent dropout~\cite{Gal16rnndropout} on RNNs ($p=0.2$).
We train the network using the widely used optimization method Adam~\cite{Kingma14adam}.
To evaluate the neural network, we use the remaining 10\% of pairs left aside for testing, 
retaining the model with the highest likelihood on the validation set.

\begin{table}[t]
\caption{Details of the NL-code data used for learning unsupervised correspondence features.}
\centering
\footnotesize
\begin{tabular}{c|c|p{1.1cm}|c|c|c|c}
\hline
\multirow{2}{*}{Lang.}  & \multirow{2}{*}{\shortstack{Training Data \\(NL/Code Pairs)}} & Validation Data & \multicolumn{2}{c|}{Intents} & \multicolumn{2}{c}{Code} \\
& & & \multicolumn{1}{p{0.9cm}}{Avg. Length} & \multicolumn{1}{p{1.1cm}|}{Vocabulary Size} & \multicolumn{1}{p{0.9cm}}{Avg. Length} & \multicolumn{1}{p{1.1cm}}{Vocabulary Size} \\
\hline
Python & 33,946 & 3,773 & 11.9 & 12,746 & 65.4 & 30,286 \\ 
Java   & 37,882 & 4,208 & 11.6 & 13,775 & 65.7 & 29,526 \\ 
\hline
\end{tabular}\vspace{-0.1cm}
\label{tab:exp:unlabeled_data}
\end{table}

%

\vspace{0.1cm}\emph{Evaluating the mining model:}
For the logistic regression classifier, which uses the structural and correspondence features 
described above, 
the latter computed by the previous neural network, we use our annotated intent/snippet data (Section~\ref{sec:annotation:dataset})\footnote{Recall that our annotated data contains only how-to style questions, and therefore question filtering is not required. When applying our mining method to the full SO data, we could use the how-to question classifier in~\cite{iyer2016summarizing}.} 
during \textbf{5-fold cross validation}.
Recall, our code mining model takes as input a SO question 
(\ie intent reflected by the question title) with its answers, and outputs a ranked list of candidate 
intent/snippet pairs (with probability scores).
For evaluation, we first rank all candidate intent/snippet pairs for all questions, and then 
compare the ranked list with gold-standard annotations. 
We present the results using standard precision-recall (PR) and Receiver Operating Characteristic 
(ROC) curves. 
In short, a PR curve shows the precision w.r.t.~recall for the top-$k$ predictions in the ranked list, 
with $k$ from 1 to the number of candidates.
A ROC curve plots the true positive rates w.r.t.~false positive rates in similar fashion.
We also compute the Area Under the Curve (AUC) scores for all ROC curves.

\def\model/{\textsc{Full}}
\def\structural/{\textsc{Structural}}
\def\correspondence/{\textsc{Correspondence}}

\def\acconly/{\textsc{AcceptOnly}}
\def\allpost/{\textsc{All}}
\def\random/{\textsc{Random}}

\emph{Baselines:} As baselines for our model (denoted as \model/), 
we implement three approaches reflecting prior work and sensible heuristics:
\begin{description}
	\item[\acconly/] is the state-of-the art from prior work~\cite{wong2013autocomment,iyer2016summarizing}; it selects the whole code snippet in the \textit{accepted} answers containing exactly one code snippet.
	\item[\allpost/] denotes the baseline method that exhaustively selects all full code blocks in the top-3 answers in a post.
	\item[\random/] is the baseline that randomly selects from all consecutive code segment candidates.
\end{description}
Similarly to our model, we enforce the constraint that all mined code snippets given by the 
baseline approaches should be parseable.

Additionally, to study the impact of hand-crafted \textbf{\structural/} versus learned 
\textbf{\correspondence/} features, we also trained versions of our model with either of 
the two types of features only.

\subsection{Results and Discussion}
\label{sec:experiments:results}


%


\begin{figure*}[t!]
	\centering
	\subfloat[ROC Curve with AUC Scores on Python]{\label{fig:exp:py:roc}{\includegraphics[width=0.4 \textwidth]{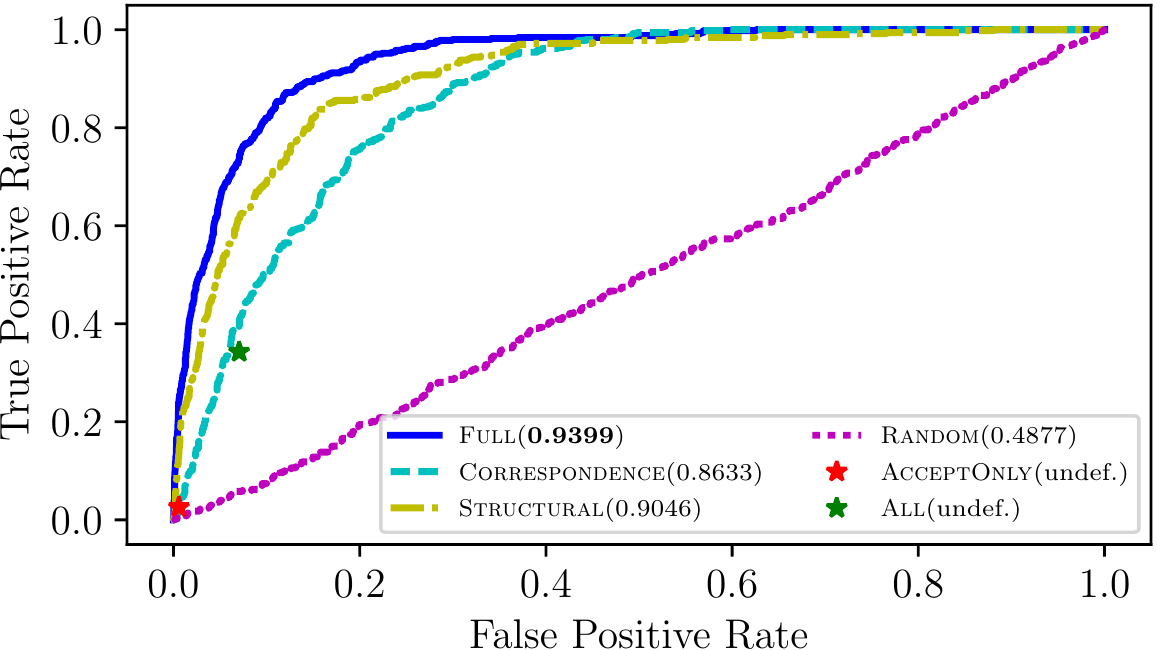} }}%
    \qquad
    \subfloat[Precision-Recall Curve on Python]{\label{fig:exp:py:pr}{\includegraphics[width=0.4 \textwidth]{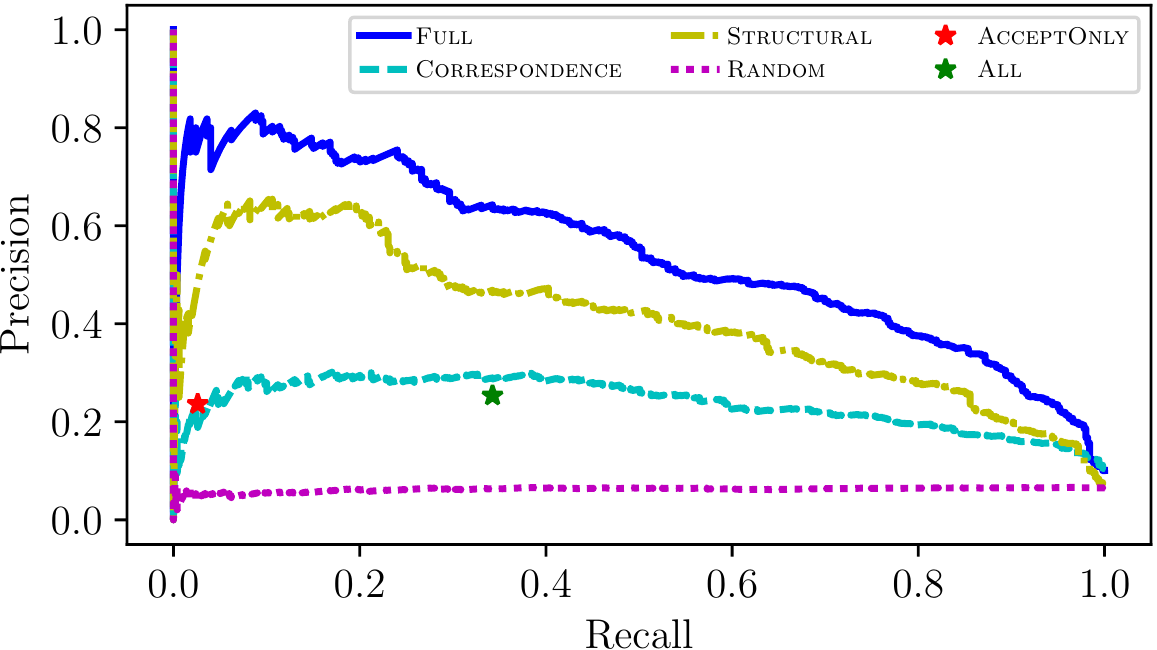} }}

%
	\subfloat[ROC Curve with AUC Scores on Java]{\label{fig:exp:java:roc}{\includegraphics[width=0.4 \textwidth]{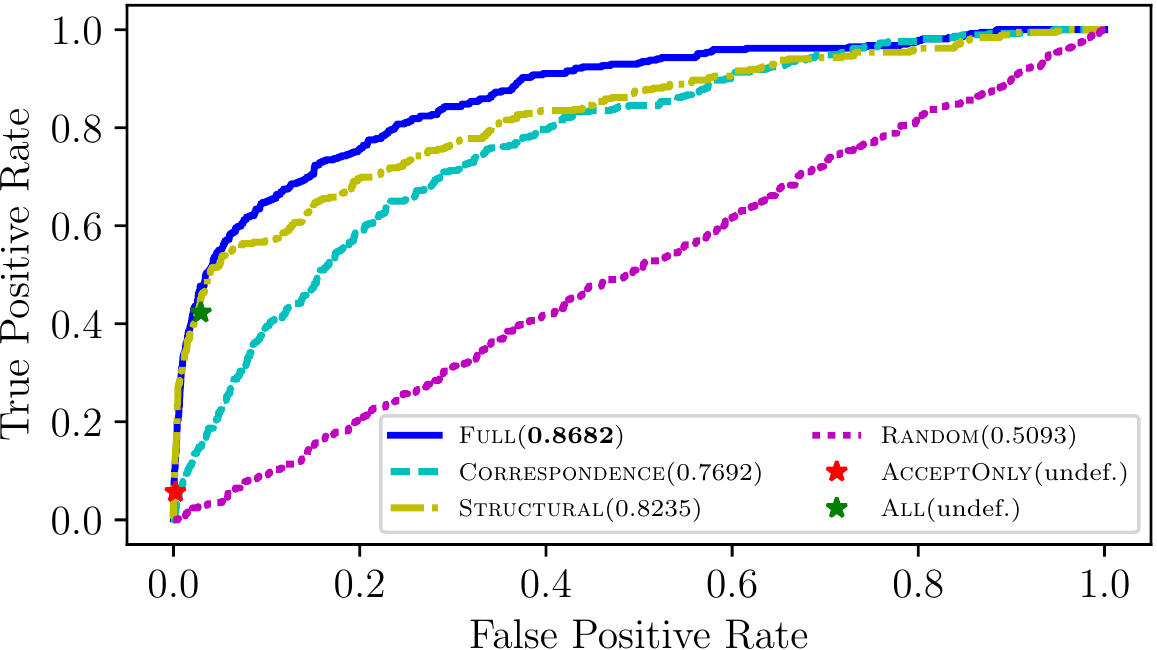} }}%
    \qquad
    \subfloat[Precision-Recall Curve on Java]{\label{fig:exp:java:pr}{\includegraphics[width=0.4 \textwidth]{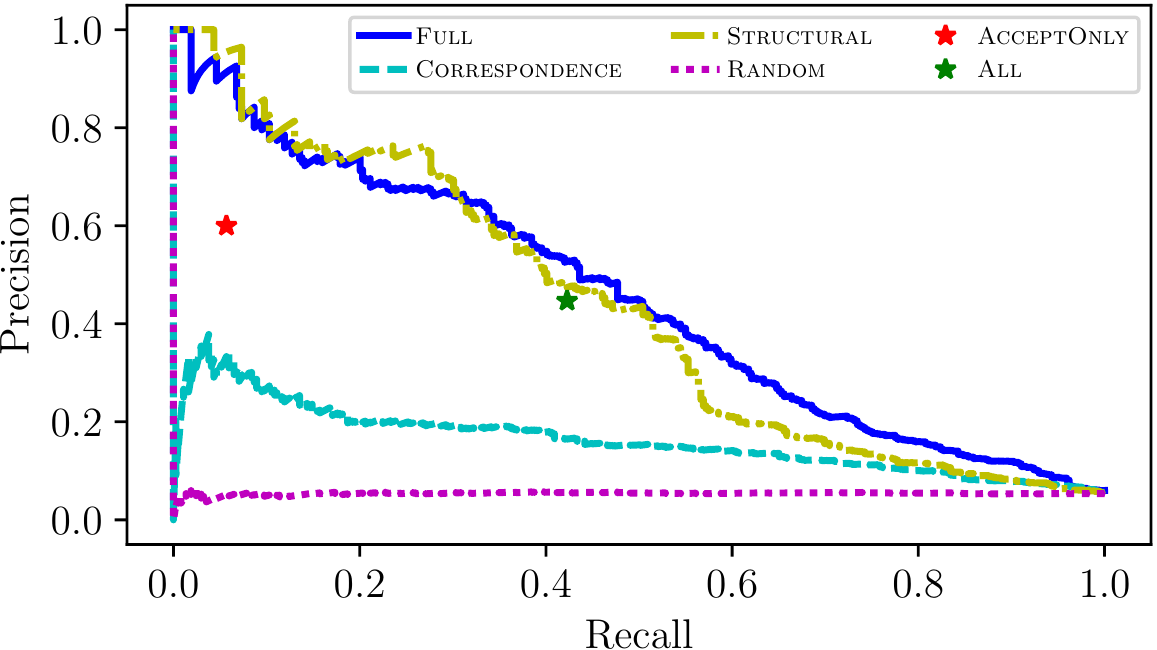} }}%
    \caption{Evaluation Results on Mining Python (a)(b) and Java (c)(d)}
	\label{fig:exp:main_results}
\end{figure*}

Our main results are depicted in Figure~\ref{fig:exp:main_results}.
First, we can see that the precision of the \textsc{random} baseline is only 0.10 for Python 
and 0.06 for Java.
This indicates that only one in 10-17 candidate code snippets is judged to validly correspond 
to the intent, reflecting the difficulty of the task.
The \acconly/ and \allpost/ baselines perform significantly better, with precision of 0.5 or 0.6 at 
recall 0.05-0.1 and 0.3-0.4 respectively, indicating that previous heuristic methods using full 
code blocks are significantly better than random, but still have a long way to go to extract 
broad-coverage and accurate NL-code pairs (particularly in the case of Python).%
\footnote{Interestingly, \acconly/ and \allpost/ have similar precision, which might be due to two facts.
First, we enforce all candidate snippets to be syntactically correct, which rules out erroneous 
candidates like input/output examples.
Second, we use the top 3 answers for each question, which usually have relatively high quality.}

Next, turning to the full system, we can see that the method with the full feature set significantly 
outperforms all baselines (Figures~\ref{fig:exp:py:pr} and~\ref{fig:exp:java:pr}):
much better recall (precision) at the same level of precision (recall) as the heuristic approaches.
The increase in precision suggests the importance of intelligently selecting NL-code pairs using 
informative features, and the increase in recall suggests the importance of considering segments 
of code within code blocks, instead of simply selecting the full code block as in prior work.

Comparing different types of features (\structural/ v.s.~\correspondence/), we find
that with structural features alone our model already significantly outperforms baseline 
approaches; and these features are particularly effective for Java.
On the other hand, interestingly the correspondence features alone provide less competitive results.
Still, the structural and correspondence features seem to be complementary, with the 
combination of the two feature sets further significantly improving performance, particularly 
on Python.
A closer examination of the results generated the following insights.

\vspace{0.1cm}\emph{Why do correspondence features underperform?}
While these features effectively filter \emph{totally} unrelated snippets, they still 
have a difficult time excluding related contextual statements, \eg imports, assignments.
This is because (1) the snippets used for training correspondence features are full code blocks (as in~\S\ref{sec:mining:correspondence}), usually starting 
with \texttt{import} statements; and (2) the library names in \texttt{import} statements often have 
strong correspondence with the intents (\eg \textit{``How to get current time in Python?'' and 
\texttt{import datetime}}), yielding high correspondence probabilities. 

\vspace{0.1cm}\emph{What are the trends and error cases for structural features?}
Like the baseline methods, \structural/ tends to give priority to full code blocks; out of the 
top-100 ranked candidates for \structural/, all were full code blocks (in contrast to only 21 
for \correspondence/).
Because selecting code blocks is a reasonably strong baseline, and because the model has 
access to other strongly-indicative binary features that can be used to further prioritize its 
choices, it is able to achieve reasonable precision-recall scores only utilizing these features.
However, unsurprisingly, it lacks fine granularity in terms of pinpointing exact code segments 
that correspond to the intents; when it tries to select partial code segments, the results are likely 
to be irrelevant to the intent.
As an example, we find that \structural/ tends to select the last line of code at each code block, 
since the learned weights for \textsc{LineNum=1} and \textsc{EndsCodeBlock} features are high, 
even though these often consist of auxiliary \texttt{print} statement or even simply \texttt{pass} 
(for Python).

\vspace{0.1cm}\emph{What is the effect of the combination?}
When combining \structural/ and \correspondence/ features together, the full model has the 
ability to use the knowledge of the \structural/ model extract full code blocks or ignore imports, 
leading to high performance in the beginning stages.
Then, in the latter and more difficult cases, it is able to more effectively cherry-pick smaller 
snippets based on their correspondence properties, which is reflected in the increased accuracy 
on the right side of the ROC and precision-recall curves.

\vspace{0.1cm}\emph{How do the trends differ between programming languages?}
Compared with the baseline approaches \acconly/ and \allpost/, our full model performs 
significantly better on Python.
We hypothesize that this is because learning correspondences between intent/snippet pairs 
for Java is more challenging. 
Empirically, Python code snippets are much shorter, and the average number of tokens for 
predicted code snippets on Python and Java is 11.6 and 42.4, respectively.
Meanwhile, since Java code snippets are more verbose and contain significantly more boilerplate 
(\eg class/function definitions, type declaration, exception handling, \etc), estimating 
correspondence scores using neural networks is more challenging. 

Also note that the \structural/ model performs much better on Java than on Python.
This is due to the fact that Java annotations are more likely to be full code blocks (see 
Table~\ref{tab:annotation:stat}), which can be easily captured by our designed features like 
\textsc{FullBlock}.
Nevertheless, adding correspondence features is clearly helpful for the harder cases for 
both programming languages.
For instance, from the ROC curve in Figure~\ref{fig:exp:java:roc}, our full model achieves 
higher true positive rates compared with \structural/, registering higher AUC scores.

\subsection{Must We Annotate Each Language?}
\label{sec:experiments:transfer_learning}


\begin{figure*}[t!]
	\centering
    \subfloat[Java $\mapsto$ Python]{\label{fig:exp:py:transfer}{\includegraphics[width=0.4 \textwidth]{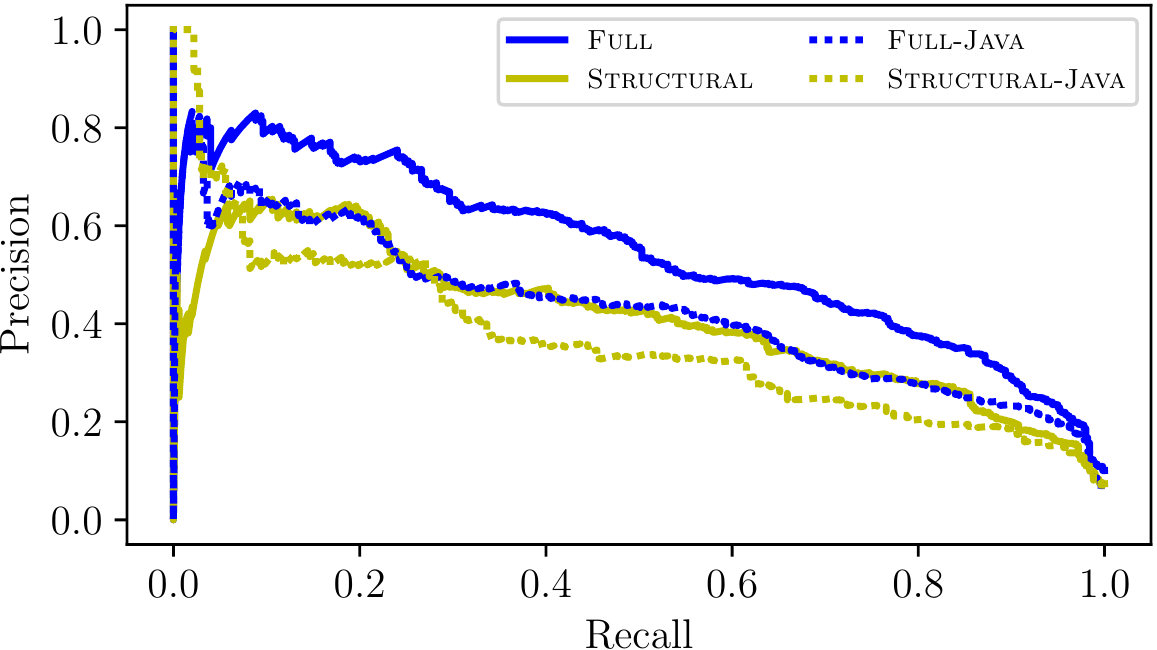} }}%
    \qquad
    \subfloat[Python $\mapsto$ Java]{\label{fig:exp:java:transfer}{\includegraphics[width=0.4 \textwidth]{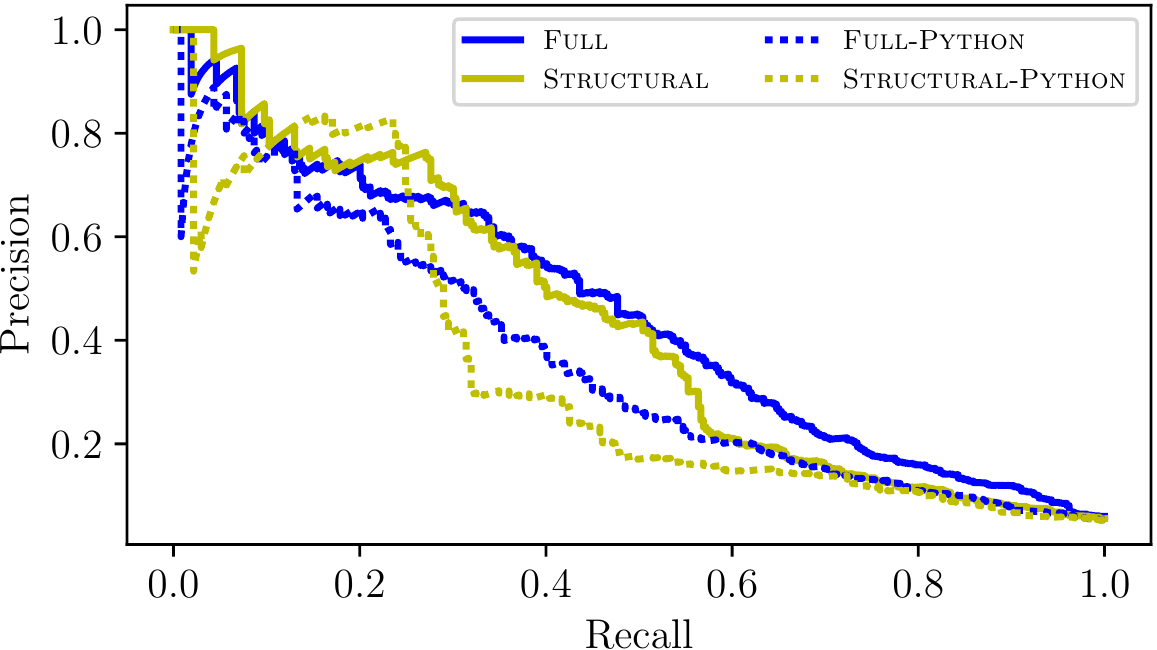} }}%
    \caption{Precision-Recall Curves for Transfer Learning on Java $\mapsto$ Python (a) and Python $\mapsto$ Java (b)}
	\label{fig:exp:transfer}
\end{figure*}

As discussed in~\S\ref{sec:annotation}, collecting high-quality intent/snippet annotations to 
train the code mining model for a programming language can be costly and time-consuming.
An intriguing research question is how we could \emph{transfer} the learned code mining model 
from one programming language and use it for mining intent/snippet data for another language.
To test this, we train a code mining model using the annotated intent/snippet data on language 
A, and evaluate using the annotated data on language B.%
\footnote{We still train the correspondence model using the target language unlabeled data.}
This is feasible since almost all of the features used in our system is language-agnostic.%
\footnote{The only one that was not applicable to both languages was the \textsc{SingleValue} 
feature for Python, which helps rule out code that contains only a single value. 
We omit this feature in the cross-lingual experiments.}
Also note values of a specific feature might have different ranges for different languages. 
As an example, the average value of \textsc{SGivenI} feature for Python and Java is -23.43 
and -47.64, respectively.
To mitigate this issue, we normalize all feature values to zero mean and unit variance before 
training the logistic regression classifier.

Figures~\ref{fig:exp:py:transfer} and~\ref{fig:exp:java:transfer} show the precision-recall curves 
for applying Java (Python) mining model on Python (Java) data.
We report results for both the \structural/ model and our full model, and compare with the original 
models trained on the target programming language.
Unsurprisingly, the original full model tuned on the target language still performs the best.
Nevertheless, we observe that the performance gap between the original full model and the 
transferred one is surprisingly small.
Notably, we find that overall the transferred full model (\textsc{Full-Java}) performs second best 
on Python, even outperforming the original \structural/ model.
These results are encouraging, in that they suggest that it is likely feasible to train a single code 
mining classifier and then apply it to different programming languages, even those for which we 
do not have any annotated intent/snippet data.

\subsection{Successful and Failed Examples}
\label{sec:experiments:case_study}

As illustration, we showcase successful and failed examples of our proposed approach, 
for Python in Table~\ref{tab:exp:examples_py} and for Java in Table~\ref{tab:exp:examples_java}.
Given a SO question (intent), we show the top-3 most probable code snippets.
First, we find our model can correctly identify code snippets for various types of intents, 
even when the target snippets are not full code blocks.
$I_1$ and $I_6$ demonstrate that our model can leave contextual information like variable 
definitions in the original SO posts and only retain the actual implementation of the intent.%
\footnote{We refer readers to the original SO page for reference.}
$I_2$, $I_3$ and $I_7$ are more interesting: in the original SO post, there could be multiple 
possible solutions in the same code block ($I_2$ and $I_7$), or the gold-standard snippets 
are located inside larger code structures like a for loop ($S_2$ for $I_3$). 
Our model learns to ``break down'' the solutions in single code block into multiple snippets, 
and extract the actual implementation from large code chunks.

We also identify four sources of errors:
\begin{itemize}
	\item \textit{Incomplete code:} Some code snippets are incomplete, and the model fails 
	to include intermediate statements (\eg definitions of custom variables or functions) that 
	are part of the implementation. For instance, $S_3$ for $I_3$ misses the definition of the
	 \texttt{keys\_to\_keep}, which is the set of keys excluding the key to remove.
	\item \textit{Including auxiliary info:} Sometimes the model fails to exclude auxiliary code 
	segments like the extra context definition (\eg $S_1$ for $I_8$) and \texttt{print} function. 
	This is especially true for Java, where full code blocks are likely to be correct snippets, 
	and the model tends to bias towards larger code chunks.
	\item \textit{Spurious cases:} 
	We identify two ``spurious'' cases where our correspondence feature often do not suffice. 
	(1) \emph{Counter examples}: the $S_1$ for $I_4$ is mentioned in the original post as a 
	counter example, but the values of correspondence features are still high since \texttt{append()} 
	is highly related to \textit{``append it to another list''} in the intent.
	(2) \emph{Related implementation}: $I_5$ shows an example where the model has 
	difficulty distinguishing between the actual snippets and related implementations.
	\item \textit{Annotation error:} We find cases where our annotation is incomplete. For 
	instance, $S_1$ for $I_9$ should be correct. As discussed in Section~\ref{sec:annotation}, 
	guaranteeing coverage in the annotation process is non-trivial, and we leave this 
	as a challenge for future work.
\end{itemize}





%

\begin{table}[tb]
  \caption{Examples of Mined Python Code\vspace{-0.2cm}}
  \label{tab:exp:examples_py}
  \centering
  \small
  \begin{tabular}{lp{7.6cm}}
  \hline

  \hline
  \multicolumn{2}{l}{{\bf $I_1$:} {\it Remove specific characters from a string in python}} \\
  \multicolumn{2}{l}{{\text URL: }\url{https://stackoverflow.com/questions/3939361/}} \\ \hdashline
  \multicolumn{2}{l}{{\bf Top Predictions:}} \\
  $S_1$ &
\begin{lstlisting}[style=python]
string.replace('1', '') |\cmark|
\end{lstlisting}
\\
  $S_2$ &
\begin{lstlisting}[style=python]
line = line.translate(None, '!@#$') |\cmark|
\end{lstlisting}
\\ 
  $S_3$ &
\begin{lstlisting}[style=python]
line = re.sub('[!@#$]', '', line) |\cmark|
\end{lstlisting}
\\ \hline
  \multicolumn{2}{l}{{\bf $I_2$:} {\it Get Last Day of the Month in Python}} \\ 
  \multicolumn{2}{l}{{\text URL: }\url{https://stackoverflow.com/questions/42950/}} \\ \hdashline
  \multicolumn{2}{l}{{\bf Top Predictions:}} \\
  $S_1$ &
\begin{lstlisting}[style=python]
calendar.monthrange(year, month)[1] |\cmark|
\end{lstlisting}
\\
  $S_2$ &
\begin{lstlisting}[style=python]
calendar.monthrange(2100, 2) |\cmark|
\end{lstlisting}
\\ 
  $S_3$ &
\begin{lstlisting}[style=python]
(datetime.date(2000, 2, 1) - datetime.timedelta(days=1))|\cmark|
\end{lstlisting}
\\ \hline
  \multicolumn{2}{l}{{\bf $I_3$:} {\it Delete a dictionary item if the key exists}} \\
  \multicolumn{2}{l}{{\text URL: }\url{https://stackoverflow.com/questions/15411107/}} \\ \hdashline
  \multicolumn{2}{l}{{\bf Top Predictions:}} \\
  $S_1$ &
\begin{lstlisting}[style=python]
mydict.pop('key', None) |\cmark|
\end{lstlisting}
\\
  $S_2$ &
\begin{lstlisting}[style=python]
del mydict[key] |\cmark|
\end{lstlisting}
\\ 
  $S_3$ &
\begin{lstlisting}[style=python]
new_dict = {k: mydict[k] for k in keys_to_keep}|\xmark|
\end{lstlisting}
\\ \hline
  \multicolumn{2}{l}{{\bf $I_4$:} {\it Python: take the content of a list and append it to another list}} \\
  \multicolumn{2}{l}{{\text URL: }\url{https://stackoverflow.com/questions/8177079/}} \\ \hdashline
  \multicolumn{2}{l}{{\bf Top Predictions:}} \\
  $S_1$ &
\begin{lstlisting}[style=python]
list2.append(list1) |\xmark|
\end{lstlisting}
\\
  $S_2$ &
\begin{lstlisting}[style=python]
list2.extend(list1) |\cmark|
\end{lstlisting}
\\ 
  $S_3$ &
\begin{lstlisting}[style=python]
list1.extend(mylog) |\cmark|
\end{lstlisting}
\\ \hline
  \multicolumn{2}{l}{{\bf $I_5$:} {\it Converting integer to string in Python?}} \\
  \multicolumn{2}{l}{{\text URL: }\url{https://stackoverflow.com/questions/961632/}} \\ \hdashline
  \multicolumn{2}{l}{{\bf Top Predictions:}} \\
  $S_1$ &
\begin{lstlisting}[style=python]
int('10') |\xmark|
\end{lstlisting}
\\
  $S_2$ &
\begin{lstlisting}[style=python]
str(10); int('10') |\xmark|
\end{lstlisting}
\\ 
  $S_3$ &
\begin{lstlisting}[style=python]
a.__str__() |\cmark|
\end{lstlisting}
\\ \hline
\end{tabular}\vspace{-0.2cm}
\end{table}

\begin{table}[tb]
  \caption{Examples of Mined Java Code\vspace{-0.2cm}}
  \label{tab:exp:examples_java}
  \centering
  \small
  \begin{tabular}{lp{7.6cm}}
  \hline

  \hline
  \multicolumn{2}{l}{{\bf $I_6$:} {\it How to convert List<Integer> to int[] in Java?}} \\ 
  \multicolumn{2}{l}{{\text URL: }\url{https://stackoverflow.com/questions/960431/}} \\ \hdashline
  \multicolumn{2}{l}{{\bf Top Predictions:}} \\
  $S_1$ &
\begin{lstlisting}[style=java]
int[] array = list.stream().mapToInt(i -> i).toArray(); |\cmark|
\end{lstlisting}
\\
  $S_2$ &
\begin{lstlisting}[style=java]
int[] intArray2 = ArrayUtils.toPrimitive(myList.toArray(NO_INTS)); |\xmark|
\end{lstlisting}
\\ 
  $S_3$ &
\begin{lstlisting}[style=java]
int[] intArray = ArrayUtils.toPrimitive(myList.toArray(new Integer[myList.size()])); |\cmark|
\end{lstlisting}
\\ \hline
  \multicolumn{2}{l}{{\bf $I_7$:} {\it How do I compare strings in Java?}} \\
  \multicolumn{2}{l}{{\text URL: }\url{https://stackoverflow.com/questions/513832/}} \\ \hdashline
  \multicolumn{2}{l}{{\bf Top Predictions:}} \\
  $S_1$ &
\begin{lstlisting}[style=java]
new String("test").equals("test"); |\cmark|
\end{lstlisting}
\\
  $S_2$ &
\begin{lstlisting}[style=java]
Objects.equals(null, "test"); |\cmark|
\end{lstlisting}
\\ 
  $S_3$ &
\begin{lstlisting}[style=java]
nullString1.equals(nullString2); |\cmark|
\end{lstlisting}
\\ \hline
  \multicolumn{2}{l}{{\bf $I_8$:} {\it How do I set the colour of a label (coloured text) in Java?}} \\
  \multicolumn{2}{l}{{\text URL: }\url{https://stackoverflow.com/questions/2966334/}} \\ \hdashline
  \multicolumn{2}{l}{{\bf Top Predictions:}} \\
  $S_1$ &
\begin{lstlisting}[style=java]
JLabel title = new JLabel("I love stackoverflow!", JLabel.CENTER);
title.setForeground(Color.white); |\xmark|
\end{lstlisting}
\\
  $S_2$ &
\begin{lstlisting}[style=java]
frame.add(new JLabel("<html>Text color: <font color='red'>red</font></html>")); |\cmark|
\end{lstlisting}
\\ 
  $S_3$ &
\begin{lstlisting}[style=java]
label.setForeground(Color.red); |\cmark|
\end{lstlisting}
\\ \hline
  \multicolumn{2}{l}{{\bf $I_9$:} {\it Generating a Random Number between 1 and 10 Java}} \\
  \multicolumn{2}{l}{{\text URL: }\url{https://stackoverflow.com/questions/20389890/}} \\ \hdashline
  \multicolumn{2}{l}{{\bf Top Prediction:}\hfill(only show one for space reason)} \\
  $S_1$ &
\begin{lstlisting}[style=java]
public static int randInt(int min, int max) {
    Random rand = new Random();
    int randomNum = rand.nextInt((max - min) + 1) + min;
    return randomNum; } |\xmark| |(annotation error)|
\end{lstlisting}
\\ \hline
\end{tabular}
\vspace{-0.3cm}
\end{table}

\section{Related Work}

A number of previous works have proposed methods for mining intent-snippet pairs
for purposes of code summarization, search, or synthesis.
We can view these methods from several perspectives:

\vspace{0.1cm}\emph{Data Sources:}
First, what data sources do they use to mine their data?
Our work falls in the line of mining intent-snippet pairs from SO (\eg\cite{zagalsky2012example, wong2013autocomment,yao18so,iyer2016summarizing}), while there has been research on mining from other data sources
such as API documentation \cite{chatterjee2009sniff, 
movshovitz2013natural,barone2017parallel}, code comments \cite{wong2015clocom}, 
specialized sites \cite{quirk2015language}, parameter/method/class names 
\cite{sridhara2011parametercomments,allamanis2015suggesting}, and 
developer mailing lists \cite{panichella2012mining}.
It is likely that it could be adapted to work with other sources, requiring
only changes in the definition of our structural features to incorporate
insights into the data source at hand.

\vspace{0.1cm}\emph{Methodologies:}
Second, what is the methodology used therein, and can it scale to our task of gathering 
large-scale data across a number of languages and domains?
Several prior work approaches used heuristics to extract aligned intent-snippet pairs%
~\cite{chatterjee2009sniff,zagalsky2012example,wong2013autocomment}).
Our approach also contains an heuristic component.
However, as evidenced by our experiments here, our method is more effective at extracting 
accurate intent-snippet pairs. 

Some work on code search has been performed by retrieving candidate code snippets given an intent based on weighted keyword 
matches and other features \cite{wei2015building,niu2016learning}.
These methods similarly aim to learn correspondences between natural language queries 
and returned code, 
but they are tailored specifically for performing code search, apply a more rudimentary
feature set (\eg they do not employ neural network-based correspondence features) than 
we do, and will generally not handle sub-code-block sized contexts, which proved 
important in our work.

We note that concurrent to this work,~\cite{yao18so} also explored the problem of mining intent/code pairs from SO, identifying candidate code blocks of an intent using information from both the contextual texts and the code in an SO answer.
Our approach, however, considers more fine-grained, sub-code-block sized candidates, aiming to recover code solutions that \emph{exactly} answer the intent. 

Finally, some work has asked programmers to manually write NL descriptions for code 
\cite{oda2015learning,lin2018nl2bash}, or vice-versa \cite{wang2015overnight}.
This allows for the generation of high-quality data, but is time consuming and 
does not scale beyond limited domains.


\section{Threats to Validity}

Besides threats related to the manual labeling (Section~\ref{sec:annotation:dataset}), 
we note the following overall threats to the validity of our approach:

\emph{Annotation Error:} Our code mining approach is based on learning from a small amount of annotated data, and errors in annotation may impact the performance of the system (see Sections~\ref{sec:annotation} and~\ref{sec:experiments:case_study}).

\emph{Data Set Volume:} Our annotated data set contains mainly high-ranked SO questions, and is relatively small (with a few hundreds of examples for each language), which could penitentially hinder the generalization ability of the system on lower-ranked questions. Meanwhile, we used cross-validation for evaluation, while evaluating our mining method on full-scale SO data would be ideal but challenging.

\section{Conclusions}

In this paper, we described a novel method for extracting aligned code/natural language pairs 
from the Q\&A website \SO.
The method is based on learning from a small number of annotated examples, using highly 
informative features that capture structural aspects of the code snippet and the correspondence 
between it and the original natural language query.
Experiments on Python and Java demonstrate that this approach allows for more accurate and 
more exhaustive extraction of NL-code pairs than prior work.
We foresee the main impact of this paper lying in the resources it would provide when applied to the full \SO data: the NL-code pairs 
extracted would likely be of higher quality and larger scale.
Given that high-quality parallel NL-code data sets are currently a significant bottleneck in the 
development of new data-driven software engineering tools,
we hope that such a resource will move the field forward.
In addition, while our method is relatively effective compared to previous work, there is still 
significant work to be done on improving mining algorithms to deal with current failure cases, 
such as those described in Section~\ref{sec:experiments:case_study}.
Our annotated data set and evaluation tools, publicly available, may provide an impetus 
towards further research in this area.


\balance
\bibliographystyle{ACM-Reference-Format}
\bibliography{main}

%
%

\end{document}